\documentclass{article}
\usepackage[T1]{fontenc}
\usepackage{iclr2027_conference,times}
\usepackage{arxiv_preprint}
\usepackage{amsmath,amssymb}
\usepackage{graphicx}
\usepackage{flafter}
\usepackage{booktabs}
\usepackage{array,longtable}
\usepackage{xurl}
\usepackage{hyperref}
\graphicspath{{figures/}}

\title{Rethinking Streaming Video Diffusion Model: \\ Context, Execution, and Training}
\author{Hongchen Zhang\\
\normalfont University of Chinese Academy of Sciences\\
\normalfont\texttt{zhanghongchenucas@gmail.com}}
\hypersetup{
  hidelinks,
  pdfauthor={Hongchen Zhang},
  pdftitle={Rethinking Streaming Video Diffusion Model: Context, Execution, and Training}
}

\begin{document}
\maketitle

\begin{abstract}
Understanding the design space of streaming video diffusion is essential to exploring its potential for generation quality and computational efficiency. We develop a unified analytical framework that relates model and sampler choices, historical conditioning, execution scheduling, and training strategies. The framework accommodates a broad family of causal context-selection policies and makes their computational dependencies and training--inference alignment explicit. Within this design space, we study three representative policies: clean, same-level, and progressive history. On the full VBench prompt set, same-level and progressive history achieve aggregate scores of 85.24 and 85.60, respectively, compared with 84.45 for the clean-history reference. Long-video comparisons further show improved subject consistency and more coherent motion with progressive history. By allowing multiple denoising nodes to be processed together, progressive-history pipelining achieves $1.57$--$2.83\times$ steady-state DiT speedups under our evaluated conditions. We additionally find that LoRA adaptation of the DMD fake-score network improves generation quality using only 2.15\% as many trainable fake-score parameters as full-parameter adaptation. Together, these findings show that fully denoised history is not a prerequisite for high-quality streaming generation and motivate the joint design of historical conditioning, execution, and training.

\end{abstract}

\section{Introduction}
\label{sec:introduction}

Streaming video generation requires a model to produce temporally coherent
content while continuously making new frames available. This requirement differs
from generating an entire clip before displaying any output: both the time to
the first output and the interval between subsequent outputs matter.
Autoregressive video diffusion models provide a natural basis for this setting,
combining incremental generation over video chunks with iterative denoising
within each chunk~\citep{ho2022videodiffusion,henschel2024streamingt2v}. Their execution is therefore governed by dependencies along
two axes: video time and denoising progress~\citep{ruhe2024rollingdiffusion}. Understanding how these dependencies
interact is important for improving streaming efficiency without compromising
generation quality.

A common design conditions each new chunk on fully denoised historical chunks.
Self Forcing~\citep{huang2025selfforcing}, for example, incorporates
autoregressive rollout into training so that the model learns from its own
generated history. This addresses a training--inference mismatch in the source
of conditioning information. However, a clean-history rollout also makes the
next chunk wait for the denoising process of its predecessor to finish.
StreamDiffusionV2~\citep{feng2026streamdiffusionv2} demonstrates how stream
batching and pipelined execution can overlap work across denoising steps and
network layers. Motivated by this execution perspective, we examine the
denoising state of historical context as a modeling choice that must be
considered together with the execution schedule.

Our central question is: \emph{which denoising state should a video chunk use
for its historical context, and how does this choice shape generation quality
and execution efficiency?} Temporal causality restricts access to future video
content, while leaving room to select among the denoising states of historical
chunks~\citep{chen2024diffusionforcing}. These choices determine both the conditioning information supplied to
the model and the dependencies governing when different chunks can advance.
Historical context thus connects the modeling and execution aspects of
streaming video generation.

We develop a unified analytical framework for streaming video generation that relates model and sampler choices, historical-context construction, execution scheduling, and training strategies. The framework accommodates a broad family of causal context-selection policies and makes explicit their relationships with attention visibility, training-state sources, and supervision objectives. Within this framework, we study three representative context policies: clean, same-level, and progressive history. We examine whether partially denoised history can provide effective conditioning for streaming generation, and how different context choices shape generation quality and inference efficiency. We also discuss empirical observations on parameter-efficient fake-score adaptation and generation quality using LoRA~\citep{hu2021lora} during Distribution Matching Distillation (DMD)~\citep{yin2024dmd}.

Same-level and progressive history outperform the clean-history reference
on the full VBench prompt set~\citep{huang2024vbench}, with qualitative examples,
including long videos, illustrating more coherent subject motion and better subject consistency.
Progressive-history pipelining achieves $1.57$--$2.83\times$
steady-state DiT speedups under our evaluation conditions.
We also find that LoRA fake-score adaptation enables faster distillation
and better generation quality using only 2.15\% as many trainable
parameters as full-parameter adaptation.

Our contributions are twofold:
\begin{itemize}
    \item \textbf{A unified framework for streaming video diffusion.}
    We characterize a broad family of causal context policies and relate
    historical-context construction to model and sampler choices, execution
    scheduling, and training strategies, clarifying their implications for
    parallelism and training--inference consistency.
    \item \textbf{A framework-guided study of quality and efficiency.}
    We study the generation quality, long-video behavior, and steady-state
    inference efficiency of representative context policies, interpreting
    their behavior through the framework, and discuss parameter-efficient
    fake-score adaptation during distillation.
\end{itemize}

\section{Related Work}
\label{sec:related}

\paragraph{Video Diffusion and Autoregressive Generation.}
Streaming video generation has advanced along two complementary directions: diffusion-based synthesis and autoregressive sequence modeling. Early video diffusion extended image denoising architectures to spatiotemporal data~\citep{ho2022videodiffusion}. Subsequent progress in latent-space modeling, large-scale training, and spatiotemporal compression enabled increasingly scalable, high-resolution text-conditioned generation~\citep{blattmann2023align,
blattmann2023stablevideo,yang2024cogvideox}. In parallel, autoregressive approaches represented videos as discrete token sequences, progressing from modeling quantized video latents to multimodal, text-conditioned generation~\citep{yan2021videogpt,
kondratyuk2024videopoet}. For streaming synthesis, combining temporal autoregression with diffusion within each video chunk brings incremental generation together with iterative refinement. Within this formulation, causal architectures and few-step distillation reduce generation latency~\citep{yin2024causvid}, while training on model-generated histories helps alleviate the mismatch between training and autoregressive inference~\citep{huang2025selfforcing}. Building on these developments, our work examines how historical context, execution scheduling, and training interact in streaming video generation.

\paragraph{Noisy Context and Pipelined Execution.}

The organization of noise levels across video frames constitutes another dimension of streaming video generation. Diffusion Forcing~\citep{chen2024diffusionforcing} trains causal sequence models with independent per-token noise levels, enabling flexible sampling
schedules and conditioning on noisy history. History-Guided Video Diffusion~\citep{song2025historyguided} extends this noise-based conditioning approach to a non-causal transformer and constructs guidance by combining scores conditioned on different history subsets and noise levels. FIFO-Diffusion~\citep{kim2024fifo} performs training-free diagonal denoising over a latent queue, using partitioning and lookahead to manage noise-level
mismatch; its partitioned design also supports multi-GPU inference. SkyReels-V2~\citep{chen2025skyreelsv2} adopts non-decreasing noise schedules for long-video generation and supports synchronous and asynchronous sampling.
At the systems level, StreamDiffusionV2~\citep{feng2026streamdiffusionv2}
combines stream batching, rolling KV caching, and pipeline orchestration across denoising steps and network layers for interactive streaming.
These approaches couple noise-state choices with different conditioning mechanisms, attention structures, and execution schemes. Our framework
separates these design elements and uses representative context policies to study their effects on generation quality
and inference efficiency.

\paragraph{Distillation and Parameter-Efficient Adaptation.}

Diffusion distillation reduces sampling cost by learning generators that require fewer denoising steps. Progressive distillation~\citep{salimans2022progressive} repeatedly halves the sampling steps of a deterministic teacher, while consistency
models~\citep{song2023consistency} learn mappings that agree along probability-flow trajectories, enabling one- and few-step generation.
Distribution Matching Distillation (DMD)~\citep{yin2024dmd} aligns student
and teacher output distributions using teacher scores and a learned fake-score model. DMD2~\citep{yin2024dmd2} links training instability to inaccurate tracking of the evolving student distribution by the fake-score model and
stabilizes optimization through a two-time-scale update rule.
Parameter-efficient adaptation offers a complementary direction: LoRA~\citep{hu2021lora} parameterizes weight updates with low-rank matrices,
and LCM-LoRA~\citep{luo2023lcmlora} applies this parameterization to latent consistency distillation of the generator. Our training-side study instead
applies low-rank adaptation to the auxiliary fake-score network in DMD, examining how its parameterization affects optimization cost and student generation quality alongside context-policy adaptation.

\section{A Unified Framework for Streaming Video Generation}
\label{sec:framework}

To place historical denoising-state selection in context, we first formulate the streaming video generation process. Our formulation describes how video is partitioned into generation units, how these units are denoised and conditioned, how computation is organized, and how the model is trained.

\subsection{Generation Process and Design Elements}
\label{sec:generation-process}

Consider a sequence of $B$ latent video chunks conditioned on $c$. The chunk
size determines the temporal granularity of generation. Each chunk follows a
noise schedule $0=t_0<t_1<\cdots<t_K$, where $x_b^j$ denotes the state of
chunk $b$ at noise level $t_j$. During generation, $x_b^K$ is the initial
noisy state and $x_b^0$ is the clean endpoint. One denoising update consists
of a network prediction followed by a sampling step:
\begin{equation}
    y_{b,j}=F_\theta(x_b^j,t_j;\mathcal C_{b,j},M_{b,j},c),
    \qquad
    x_b^{j-1}=S_j(x_b^j,y_{b,j};\xi_{b,j}).
    \label{eq:framework-update}
\end{equation}
Here $F_\theta$ is the denoising network, whose output may parameterize noise,
score, clean-sample, or velocity prediction; $S_j$ is the corresponding
sampling update, with optional randomness $\xi_{b,j}$. The noise schedule,
number of updates, and sampler govern progression within each chunk.

The context $\mathcal C_{b,j}$ and attention mask $M_{b,j}$ govern interaction between chunks. Constructing them involves selecting historical states and maintaining their cached representations. The execution pattern determines which chunks can advance together. Training further specifies the sources of input states, the supervision signal, and the parameters and gradient paths being optimized. These are distinct design choices within the same generation process.

\subsection{Historical Context and Attention}
\label{sec:historical-context}

\begin{figure*}[t]
    \centering
    \includegraphics[width=0.8\textwidth]{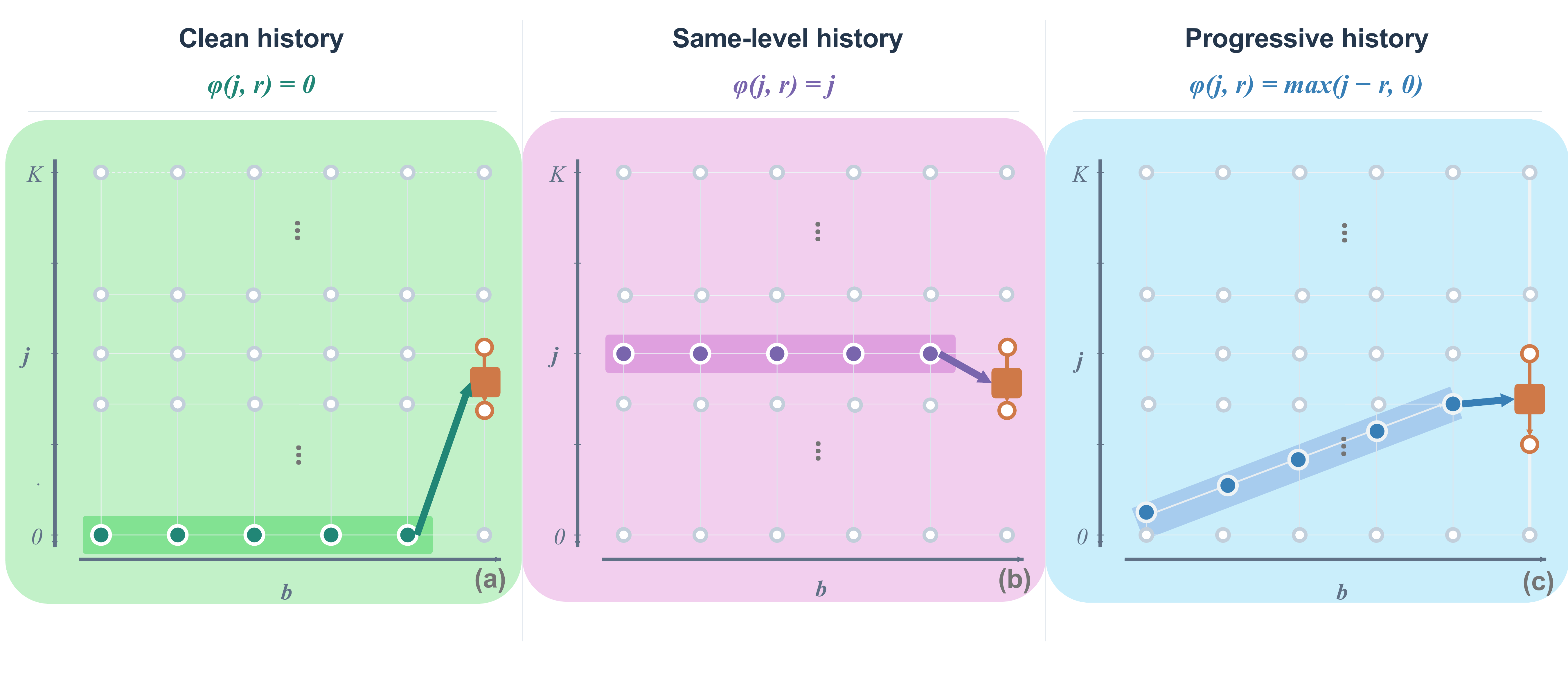}
    \caption{Representative historical context policies. Columns correspond
    to video chunks and rows to denoising levels. Highlighted historical states
    condition the current denoising update shown in orange: (a) clean history,
    (b) same-level history, and (c) progressive history.}
    \label{fig:context-policies}
\end{figure*}

\paragraph{Historical state selection.}
For a history window of $W$ chunks, let
$\mathcal R_{b,j}\subseteq\{1,\ldots,\min(W,b-1)\}$ contain the selected
history offsets. A policy $\Phi_b(j,r)\in\{0,\ldots,K\}$ specifies the noise
state selected from chunk $b-r$:
\begin{equation}
    \mathcal H_{b,j}^{\Phi}
    =\bigl\{x_{b-r}^{\Phi_b(j,r)}:r\in\mathcal R_{b,j}\bigr\}.
    \label{eq:selected-history}
\end{equation}
The selection may depend on chunk position and causally available information.
For position-independent instances, we use the shorter notation $\phi(j,r)$.
Clean, same-level, and progressive history are three representative choices:
\begin{equation}
    \phi_{\mathrm{clean}}(j,r)=0,\qquad
    \phi_{\mathrm{same}}(j,r)=j,\qquad
    \phi_{\mathrm{prog}}(j,r)=\max(j-r,0).
    \label{eq:three-policies}
\end{equation}
They respectively select clean endpoints, states at the current noise level,
and states that become cleaner with historical distance. Figure~\ref{fig:context-policies}
visualizes these representative instances of the general selection rule.

\paragraph{Effective context and visibility.}
$\mathcal H_{b,j}^{\Phi}$ specifies historical latent states, whereas
$\mathcal C_{b,j}$ contains the representations actually used by the denoising
network. These representations are computed from the selected latent states,
typically stored as KV caches, and read from those caches during denoising. The
mask $M_{b,j}$ defines token-level visibility:
\begin{equation}
    M_{b,j}(u,v)=
    \begin{cases}
        0, & \text{query token }u\text{ may attend to key token }v,\\
        -\infty, & \text{otherwise}.
    \end{cases}
    \label{eq:attention-visibility}
\end{equation}
Here $u$ belongs to the current chunk and $v$ to its history or current-chunk
features. The mask is added to attention logits before softmax: zero preserves
the connection, while $-\infty$ gives it zero attention weight. Thus $\phi$
selects the noise level of a historical state, while $M$ determines which
tokens from the resulting representation each query can access. This notation
supports dense causal attention as well as query-dependent sparse access within
the causal history. In our chunk-causal setting, future chunks are masked,
while within-chunk attention may be bidirectional. Historical features are
constructed under the same temporal restriction.

\subsection{Fully Serial and Pipelined Execution}
\label{sec:pipeline-execution}

Forward inference can proceed strictly one chunk at a time or overlap the
denoising of multiple chunks. The key distinction is whether a new chunk must
wait for its predecessor to complete the full denoising trajectory.

\paragraph{Fully serial execution.}
The model performs all $K$ denoising updates for chunk $b$ before starting
chunk $b+1$:
\begin{equation}
    x_1^K\rightarrow\cdots\rightarrow x_1^0
    \rightarrow x_2^K\rightarrow\cdots\rightarrow x_2^0\rightarrow\cdots.
    \label{eq:serial-execution}
\end{equation}
Generating $B$ chunks therefore requires $BK$ sequential chunk-level updates.
The first output requires $K$ updates, and successive outputs are separated by
another $K$ updates. Apart from stored history, only one chunk is actively
denoised, so denoising work does not overlap across chunks.

\begin{figure*}[t]
    \centering
    \includegraphics[width=0.8\textwidth]{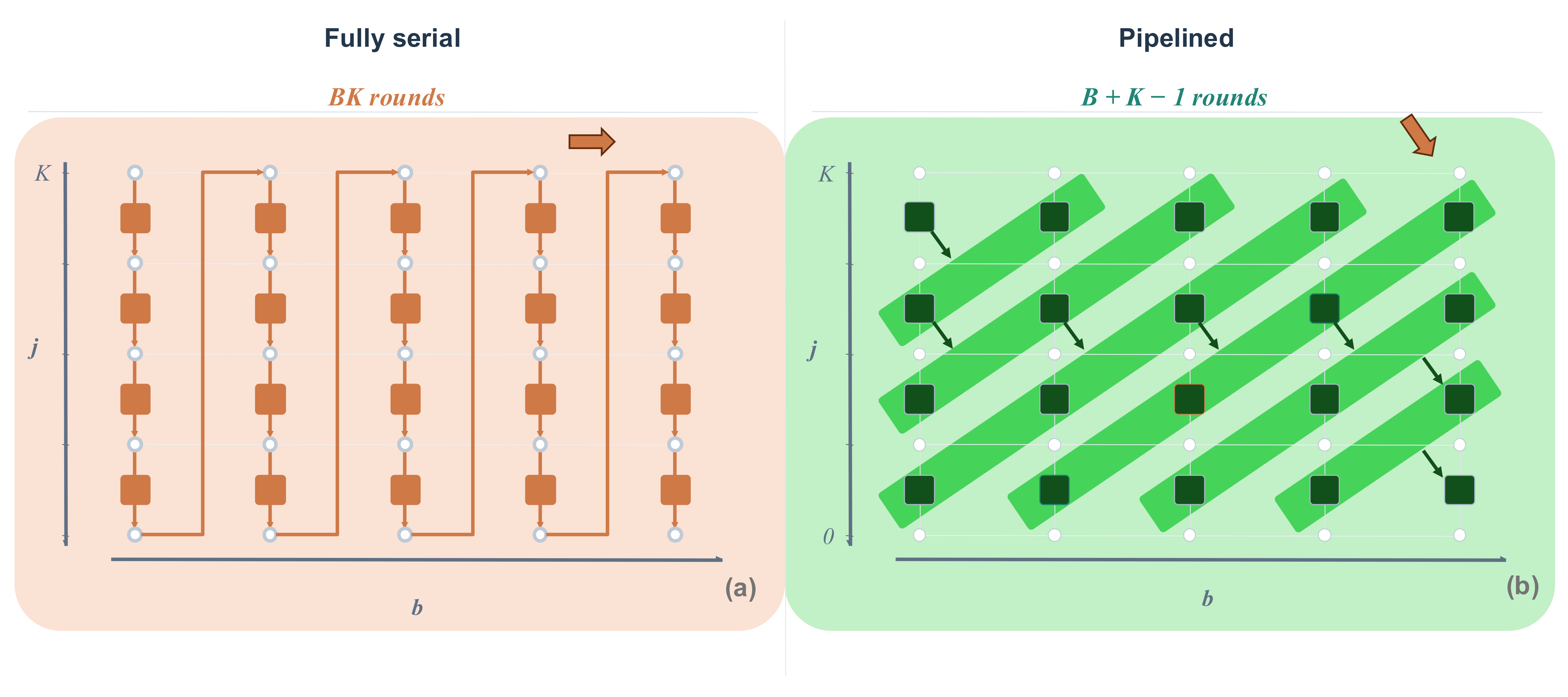}
    \caption{Fully serial and pipelined denoising. (a) Serial execution
    completes all $K$ updates of one chunk before starting the next, yielding
    $BK$ sequential rounds. (b) Pipelined execution advances multiple chunks
    at staggered denoising levels, yielding $B+K-1$ rounds under sufficient
    parallel resources while preserving the same $BK$ chunk-level updates.}
    \label{fig:execution-schedules}
\end{figure*}

\paragraph{Pipelined execution.}
Multiple chunks occupy different denoising stages, allowing their forward
computations to be grouped or overlapped. A new chunk may begin before the
preceding chunk completes denoising. For a pipeline that admits one new chunk
per round and advances every active chunk by one noise level, the active states
at the start of round $n$ are
\begin{equation}
    \bigl[x_n^K,\;x_{n-1}^{K-1},\;\ldots,\;x_{n-K+1}^{1}\bigr],
    \label{eq:pipeline-queue}
\end{equation}
with chunk indices outside $1,\ldots,B$ omitted during pipeline filling and
draining. Under sufficient concurrency, including timely construction of the
required context features, this schedule spans $K+B-1$ sequential rounds. It
still performs $BK$ chunk-level denoising updates: the reduction is in the
number of sequential rounds rather than the number of updates per chunk. The
first output traverses $K$ rounds, while a filled pipeline produces one clean
chunk per round. Higher concurrency can improve throughput but requires more
active states and intermediate features. Figure~\ref{fig:execution-schedules}
contrasts the two execution patterns.

\paragraph{Restrictions on context selection.}
Causality restricts cross-chunk conditioning to the past. In addition, a
historical state selected by the policy must already have been reached, and its
features must be available when read. Temporal causality alone does not require
clean history; the execution pattern determines how far each historical chunk
has progressed and therefore constrains the possible state choices.

Under fully serial execution, earlier chunks have completed their entire
denoising trajectories. Clean history is therefore available, and retaining
the required intermediate states and representations also permits same-level
or progressive history. Fully serial execution is not limited to clean history.

In the pipeline of Eq.~\eqref{eq:pipeline-queue}, when chunk $b$ is at level
$j$, chunk $b-r$ has reached level $\max(j-r,0)$. For states generated by the
current rollout, assuming earlier states are retained, latent availability
requires
\begin{equation}
    \Phi_b(j,r)\geq\max(j-r,0)
    \quad\text{for every historical offset actually read}.
    \label{eq:pipeline-context-constraint}
\end{equation}
For example, with $K=3$, round two processes $x_2^3$ alongside $x_1^2$.
Chunk two can select the retained level-$3$ state of chunk one under the
same-level policy, or its current level-$2$ state under the progressive policy,
but it cannot yet select the clean endpoint. 

Same-level features can be reused from earlier rounds. Progressive selection
may require features constructed in the current round, which can be supplied by
a joint causal forward pass or layer-wise execution that respects the same
visibility and feature dependencies. Thus both the selected latent states and
their effective representations must be available.

\subsection{Training Policies, State Sources, and Supervision}
\label{sec:training-consistency}

\paragraph{Policy alignment.}
When adapting a model to a specified inference process, the primary context
alignment target is
\begin{equation}
    \Phi^{\mathrm{train}}=\Phi^{\mathrm{infer}},
    \label{eq:context-alignment}
\end{equation}
together with consistent effective attention visibility and context
representation. For example, adaptation to same-level inference uses same-level
history during training rather than only clean history. If training covers a
broader family of policies, that family should include the target inference
policy. 

\paragraph{Data-derived and model-generated history.}
Historical states can be constructed directly from ground-truth data,
using either clean or noised tokens
\citep[TF and DF;][]{chen2024diffusionforcing,huang2025selfforcing},
or produced through the model's own generation process
\citep[SF and Rolling Forcing;][]{huang2025selfforcing,liu2025rollingforcing}.
Between these two constructions, partial denoising of corrupted
ground-truth chunks introduces model-generated deviations while
retaining an anchor to real data
\citep[Resampling Forcing;][]{guo2025resamplingforcing}.
The source of history therefore matters: matching historical noise
levels alone does not ensure matching state distributions or
dependencies on the current chunk.

\paragraph{Supervision objectives.}
Context construction and supervision are separate design choices. Using the
prediction $y_{b,j}$ defined above, common objectives can be summarized in
three forms. Local regression uses
\begin{equation}
    \mathcal L_{\mathrm{local}}
    =\mathbb E\!\left[
    \frac{\sum_{b,j}a_{b,j}w_{b,j}
    \lVert y_{b,j}-y_{b,j}^{\star}\rVert_2^2}
    {\sum_{b,j}a_{b,j}}
    \right],
    \label{eq:local-supervision}
\end{equation}
where $y_{b,j}^{\star}$ is the analytical target for the chosen prediction
parameterization, $a_{b,j}\in\{0,1\}$ marks supervised nodes, and $w_{b,j}$ is
a loss weight. Teacher prediction and trajectory matching use
\begin{align}
    \mathcal L_{\mathrm{teacher}}
    &=\mathbb E\!\left[
    \frac{\sum_{b,j}a_{b,j}w_{b,j}
    \lVert y_{b,j}-y_{b,j}^{T}\rVert_2^2}
    {\sum_{b,j}a_{b,j}}
    \right],\label{eq:teacher-supervision}\\
    \mathcal L_{\mathrm{trajectory}}
    &=\mathbb E\!\left[
    \left\lVert S_j(x_b^j,y_{b,j};\xi_{b,j})-x_{b,T}^{j-1}
    \right\rVert_2^2
    \right],\label{eq:trajectory-supervision}
\end{align}
where $y_{b,j}^{T}$ is a teacher prediction in the student's parameterization
and $x_{b,T}^{j-1}$ is the state obtained by integrating the teacher from
$x_b^j$ to $t_{j-1}$. Finally, video-level distribution matching can be written
as
\begin{align}
    \mathcal L_{\mathrm{distribution}}
    &=\mathbb E_c\!\left[
    D\!\left(p_{\theta}^{\mathrm{rollout}}(\cdot\mid c),
    p_{\mathrm{target}}(\cdot\mid c)\right)
    \right],\label{eq:distribution-supervision}\\
    \mathcal L_{\mathrm{DMD}}
    &=\mathbb E_{c,t}\!\left[
    D_{\mathrm{KL}}\!\left(
    p_{\theta,t}^{\mathrm{rollout}}(\cdot\mid c)
    \,\middle\|\,
    p_{\mathrm{target},t}(\cdot\mid c)
    \right)
    \right].\label{eq:dmd-supervision}
\end{align}
Here $p_\theta^{\mathrm{rollout}}$ is the joint distribution of generated video
$x_{1:B}^{0}$, $p_{\mathrm{target}}$ is the data or teacher distribution, and
$D$ is a distribution discrepancy. The subscript $t$ denotes applying the same
forward noising process to both distributions; DMD uses a reverse-KL objective
without paired samples~\citep{yin2024dmd}. These objectives can be combined
with different state sources at different stages of training.

\paragraph{Training--inference consistency.}
At the distributional level, training should reflect the joint
distribution of current denoising states and historical context
encountered during inference. Training on model-generated history
exposes the model to its own prediction errors and their propagation
across chunks, which can help mitigate error accumulation during
autoregressive generation. Training and inference need not use
identical batching or execution schedules, provided that conditional
dependencies and effective context representations are preserved.

In practice, early training often begins with data-derived history such as TF
or DF even though inference uses model-generated history. At initialization,
the model's output distribution can be far from the target distribution.
Local regression, teacher prediction, or trajectory matching on heavily
corrupted model history can then emphasize the undesirable pattern of erroneous
history paired with a clean current target, while video-level DMD may provide a
weak signal when the two distributions are too far apart. An intuitive analogy
is that TF and DF provide instruction along valid solution paths, whereas SF-like training evaluates and corrects the model on its own work. They serve different stages of learning rather than opposing objectives. A streaming video training recipe therefore jointly specifies the context policy, state source, and supervision objective.

In summary, the framework relates model and sampler choices, historical
information, attention and context representations, execution, and training.
Within this broader design space, our empirical study focuses on historical
denoising-state selection through three representative policies, examining
generation quality and inference efficiency, complemented by the training-side
study of fake-score adaptation.

\section{Training}
\label{sec:training}

Within the design space developed in Section~\ref{sec:framework}, we fix
the backbone, video chunking, and sampling schedule, and compare clean,
same-level, and progressive history in generation quality, long-video
behavior, and steady-state inference efficiency. All three policies use
rank-64 LoRA adaptation of the fake-score network; an additional
full-parameter variant under progressive history examines the effect
of the parameter update scope.

\paragraph{Context-Policy Adaptation.}
\label{sec:context-policy-adaptation}

All three policies initialize the student from the causal
Wan2.1-T2V-1.3B TF-sCM checkpoint~\citep{wan2025wan} released by
Causal-rCM~\citep{zheng2026causalrcm} and use the same text training data
and DMD training setup, differing in the target context policy.
TF-sCM adapts the continuous-time consistency distillation objective
of sCM~\citep{lu2025simplifying} to teacher forcing, conditioning each
chunk on clean ground-truth history. The frozen distillation teacher
is the bidirectional Wan2.1-T2V-14B model, and the fake-score network
is initialized from the same 14B checkpoint.

During training, the student rolls out from noise under text conditioning,
generating and reading history according to the target context policy.
Each rollout represents an approximately five-second video at
$832\times480$ resolution and 16 FPS (81 frames). The Wan2.1 VAE applies
$4\times$ temporal compression, yielding 21 latent frames organized into
seven chunks of three latent frames each. All three policies
use a fixed four-step denoising schedule with random early
exit: each rollout uniformly selects one of the four predictions as
its exit point, and the resulting clean-state predictions form the
generated video.

The student is updated with a video-level DMD loss, while the fake-score
network is updated with a velocity-prediction regression loss on generated
samples. Gradients propagate only through the final denoising prediction
of each chunk, whose forward computation is recomputed by
replay~\citep{yin2024dmd,huang2025selfforcing,zheng2026causalrcm}.

All three policies use AdamW with BF16 mixed precision
and a global effective batch size of 16. The student and fake-score
learning rates are $2\times10^{-6}$ and $4\times10^{-7}$, respectively,
with one student update per five fake-score updates. We maintain an
exponential moving average (EMA) of student parameters and use the EMA
weights for evaluation. Denoising time points, teacher CFG, DMD time
sampling, and the remaining hyperparameters are provided in
Appendix~\ref{app:training}.

\paragraph{Fake-Score Parameterization.}
\label{sec:fake-score-adaptation}

The three-policy comparison uses rank-64 fake-score
LoRA~\citep{hu2021lora}, with full-parameter student updates throughout.
For all three policies, the pretrained fake-score
backbone is frozen, and LoRA branches are added to the linear projections
in its self-attention, cross-attention, and feed-forward layers,
with alpha set to 64.

Under progressive history, we additionally train a full-parameter
fake-score variant without LoRA, keeping the student initialization,
text data, denoising schedule, learning rates, and update ratio unchanged.
This comparison examines how the fake-score parameterization affects
the generation quality attained during distillation.

\section{Experimental Results}
\label{sec:experiments}

We compare clean, same-level, and progressive history to assess how
historical noise levels affect generation quality, including motion,
subject consistency, and semantic content. We also benchmark the
steady-state inference efficiency of serial and pipelined execution.

We evaluate all 16 VBench dimensions~\citep{huang2024vbench} on the full
prompt set, using matched prompts, seed indices, and four-step sampling
schedules across models. Each video
contains 81 frames at $832\times480$ resolution and 16 FPS
(approximately five seconds). Video chunking and training settings
follow Section~\ref{sec:training}; other information is provided in Appendix~B. All scores are reported on a
100-point scale, with Total, Quality, and Semantic computed using the
evaluation pipeline's normalized, weighted aggregation.

The clean-history reference is trained using the same setup as the other
two policies in Section~\ref{sec:training} and uses clean-context refresh
during inference. For each adaptation route, we report the checkpoint with the
highest Total score among its completed evaluations, comparing
the best quality attained by the available models.

\paragraph{Quantitative Results.}
\label{sec:quality-results}

\begin{table}[!htbp]
    \centering
    \caption{Generation quality under three context policies. Higher is better.}
    \label{tab:context-quality}
    \begin{tabular}{@{}lrrr@{}}
        \toprule
        Context policy & Total & Quality & Semantic \\
        \midrule
        Clean: $\phi(j,r)=0$ & 84.446 & 84.862 & 82.783 \\
        Same-level: $\phi(j,r)=j$ & 85.240 & 85.541 & 84.038 \\
        Progressive: $\phi(j,r)=\max(j-r,0)$ & \textbf{85.599} & \textbf{85.961} & \textbf{84.149} \\
        \bottomrule
    \end{tabular}
\end{table}

Table~\ref{tab:context-quality} shows that same-level and progressive history
exceed the clean-history reference by 0.794 and 1.15 points, respectively,
with improvements in both quality and semantic scores. Progressive history
achieves the highest aggregate score among these configurations. Thus,
conditioning on incompletely denoised history does not necessarily compromise
generation quality: intermediate states can also support effective streaming
generation.

The complete dimension-level results are provided in
Appendix~\ref{app:dimension-results}.
Both non-clean configurations substantially improve dynamic degree and
multiple-object scores, while also exceeding the reference in subject
consistency, aesthetic quality, and imaging quality. The observed improvement
therefore combines greater motion activity with gains in several content and
appearance dimensions, rather than simply increasing visual change.

The gains do not extend to every dimension. Both configurations score slightly
below the reference in temporal flickering and appearance style. The clearest advantages in the
current results concern dynamic degree, subject consistency, and multiple-object
content, rather than a uniform improvement in all temporal metrics.

\paragraph{Qualitative Results and Limitations.}
\label{sec:qualitative-results}

\begin{figure}[!htbp]
    \centering
    \includegraphics[width=\textwidth]{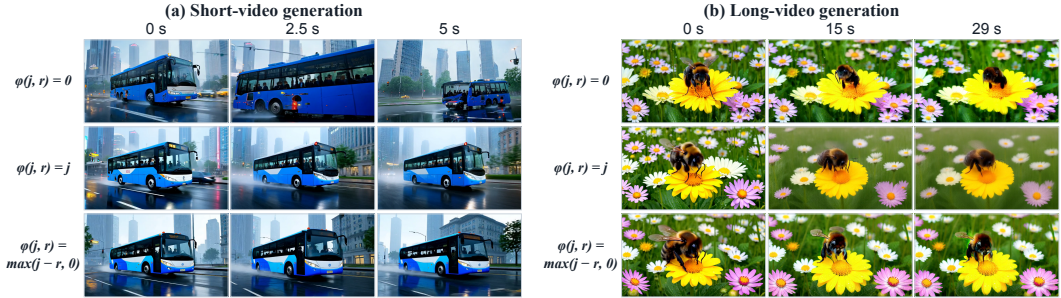}
    \caption{Three-policy comparisons under matched conditions and
    policy-specific weights. (a) Short-video example for
    \emph{``a bus accelerating to gain speed''} (seed 0).
    (b) Long-video bee example (seed 1). Columns show the indicated times.}
    \label{fig:qualitative-comparison}
\end{figure}

We further compare the visual quality of videos generated under matched
conditions. In our qualitative comparisons, clean history tends to produce
the most stable backgrounds, but moving subjects often exhibit implausible
behavior, such as overlapping limbs or objects interpenetrating their
surroundings. Progressive history produces more plausible subject motion;
compared with same-level history, it also exhibits fewer abrupt visual changes
and more stable backgrounds. Figure~\ref{fig:qualitative-comparison}(a) illustrates
vehicle interpenetration in the clean-history result, whereas the other two
examples retain a more intact bus structure at the displayed times.

We hypothesize that these differences reflect how strongly generation relies
on preceding frames. Fully denoised history provides detailed information,
which may encourage the model during training to reproduce previous
appearances rather than adapt them to physically plausible motion. Noisy
history provides less precise information about past frames, requiring the
model to rely more on learned motion and scene priors when generating the
current frame. This weaker dependence may support more plausible motion but also
reduce background stability, particularly in the initial frames
of generation.

\paragraph{Long-video generation.}
We compare approximately 30-second videos using matched prompts and
noise seeds, with each policy's corresponding trained weights.
In our qualitative observations, clean-history refresh tends to exhibit
subject-shape inconsistency and unnatural motion, while backgrounds
often remain nearly frozen, resembling static textures despite their
high temporal consistency. Progressive history better preserves subject
consistency and supports more plausible subject and background motion.
Slight background discontinuities can occur during its initial frames
but subside as generation proceeds, possibly because early conditioning
is dominated by noisy historical states before cleaner history accumulates.
Figure~\ref{fig:qualitative-comparison}(b) illustrates the contrast in subject
morphology: clean-history refresh exhibits inconsistent changes in
the bee's body shape, whereas progressive history better preserves
its form. In the full videos, clean-history refresh shows little
clear foraging-like interaction in the latter half, while progressive
history maintains more coherent flower interaction and more natural
movement of the surrounding vegetation.

\paragraph{LoRA versus Full-Parameter Fake-Score Adaptation}
\label{sec:lora-comparison}

Under progressive history, we compare LoRA and full-parameter adaptation of
the fake-score network using the training configurations in
Section~\ref{sec:fake-score-adaptation}. The student remains fully trainable
in both cases.

\begin{table}[!htb]
    \centering
    \caption{Fake-score adaptation under progressive history. }
    \label{tab:lora-comparison}
    \begin{tabular}{@{}lrrrr@{}}
        \toprule
        Fake-score adaptation & Trainable params. & Total $\uparrow$ & Quality $\uparrow$ & Semantic $\uparrow$ \\
        \midrule
        Full-parameter & 14.288B & 84.126 & 84.477 & 82.720 \\
        LoRA & 0.307B & \textbf{85.599} & \textbf{85.961} & \textbf{84.149} \\
        \bottomrule
    \end{tabular}
\end{table}

As shown in Table~\ref{tab:lora-comparison}, LoRA updates approximately
2.15\% as many fake-score parameters as full-parameter adaptation while
improving the total VBench score by 1.473 points, with gains in both quality
and semantic scores. Thus, greater adaptation freedom in the fake-score
network does not necessarily yield better student generation quality under
the current distillation setup. We hypothesize that low-rank updates impose
a structural constraint on adaptation while leveraging pretrained denoising
capabilities. This constraint may offer a more favorable balance between
adaptation and stability when tracking the evolving student distribution,
potentially providing more effective distillation signals to the student.

\paragraph{Inference Efficiency.}
We benchmark clean-history serial execution and progressive-history
pipelined execution on 4, 8, and 12 Ascend 910B2C devices using context
parallelism (CP). Each run generates 477 frames at $832\times480$
resolution and 16 FPS, with four denoising steps and one clean-state
KV refresh per chunk. We measure 20 steady-state output chunks and
report the median over three trials, including communication and cache
operations but excluding VAE decoding and pipeline fill and drain.

\begin{table}[!htbp]
    \centering
    \caption{Steady-state DiT latency (ms/chunk) and speedup.}
    \label{tab:ascend-inference}
    \begin{tabular}{@{}lrrr@{}}
        \toprule
        Devices (CP) & \shortstack[r]{Clean serial\\(ms/chunk)} &
        \shortstack[r]{Progressive pipeline\\(ms/chunk)} & Speedup \\
        \midrule
        4  & 527.7 & 336.0 & $1.57\times$ \\
        8  & 441.7 & 222.4 & $1.99\times$ \\
        12 & 626.3 & 221.0 & $2.83\times$ \\
        \bottomrule
    \end{tabular}
\end{table}

Progressive history combines multiple denoising nodes in one forward
pass, achieving $1.57$--$2.83\times$ steady-state speedups under the
evaluated configurations. As CP reduces the per-device workload,
this grouping can help amortize execution and communication overheads.
The larger speedup at CP12 mainly reflects increased serial latency,
while pipelined latency remains close to CP8. 

\section{Conclusion}
\label{sec:conclusion}
We studied streaming video diffusion as a joint design problem over historical
context, execution schedule, and training. Our framework separates four elements
that are coupled in existing systems: the policy $\Phi$ selecting the denoising
state of historical chunks, the attention mask $M$ governing visibility, the
source of historical states, and the supervision objective. This separation makes
the comparison of context policies well defined, since it identifies exactly what
must be held fixed to attribute an observed difference to $\Phi$ alone.

Within this framework we compared clean, same-level, and progressive history under
a fixed backbone, training data, chunking, denoising schedule, and DMD objective.
Both non-clean policies exceed the clean-history reference on VBench
($85.24$ and $85.60$ versus $84.45$), with progressive history additionally showing
more coherent motion and better subject consistency in long-video comparisons.
Fully denoised history is therefore not a prerequisite for high-quality streaming
generation. Because progressive history removes the dependency that forces a chunk
to wait for its predecessor to finish denoising, it admits pipelined execution, which
reduces the number of communication rounds per output chunk and yields
$1.57$--$2.83\times$ steady-state DiT speedups under our evaluated configurations.
On the training side, LoRA adaptation of the DMD fake-score network improves
generation quality while updating $2.15\%$ as many parameters as full-parameter
adaptation. Together these results argue for treating historical conditioning,
execution scheduling, and training as one design space rather than three
independent choices.

\bibliographystyle{iclr2027_conference}
\bibliography{references}

\clearpage
\appendix
\section{Training Configurations}
\label{app:training}

Table~\ref{tab:training-configurations} summarizes the configurations used
to adapt the TF-sCM-initialized student to clean, same-level, and progressive
history, together with the additional full-parameter fake-score variant
under progressive history.

\begingroup
\small
\setlength{\tabcolsep}{5pt}
\renewcommand{\arraystretch}{1.12}
\begin{longtable}{@{}>{\raggedright\arraybackslash}p{0.30\textwidth}
                       >{\raggedright\arraybackslash}p{\dimexpr0.70\textwidth-10pt\relax}@{}}
\caption{Shared training configurations for clean, same-level, and progressive
context adaptation with fake-score LoRA. The additional full-parameter
fake-score variant is evaluated under progressive history only.
The student is fully trainable in all configurations.}
\label{tab:training-configurations}\\
\toprule
Item & Setting \\
\midrule
\endfirsthead
\multicolumn{2}{l}{\tablename~\thetable\ (continued)}\\
\toprule
Item & Setting \\
\midrule
\endhead
\midrule
\multicolumn{2}{r}{Continued on the next page}\\
\endfoot
\bottomrule
\endlastfoot
Student initialization & Causal-rCM; Wan2.1-T2V-1.3B; TF-sCM;
\texttt{c3-3}. \\
Teacher / fake-score backbone & Wan2.1-T2V-14B; frozen teacher. \\
Training samples & Student-generated videos; 81 frames at 16 FPS,
approximately five seconds. \\
VAE temporal compression & Wan2.1 VAE; $4\times$ temporal compression;
81 video frames correspond to 21 latent frames. \\
Video chunking & Three latent frames per chunk; seven chunks. \\
Spatial resolution & $832\times480$. \\
Denoising schedule / random exit & Input times
$1,\ 15/16,\ 5/6,\ 5/8$; uniformly sampled exit at prediction 1--4. \\
DMD supervision & The complete generated sequence. \\
Teacher CFG / DMD time sampling & 5 / UniformShift with shift 5. \\
Gradient scope / backward computation & Stop-gradient historical caches;
replay the final denoising prediction of each chunk. \\
Optimizer implementation & NPU Master AdamW
(\texttt{npumasteradamw}). \\
Student / fake-score learning rate & $2\times10^{-6}$ / $4\times10^{-7}$. \\
AdamW parameters & Betas $(0,0.999)$; weight decay 0.01;
epsilon $10^{-8}$. \\
Student : fake-score updates & $1:5$. \\
Microbatch / gradient accumulation / effective batch size &
1 per data-parallel replica / 1 / 16 globally. \\
Training hardware and parallelism & 16 Ascend 910B2C devices;
data-parallel degree 16; context-, tensor-, and pipeline-parallel degrees 1. \\
Precision / training seed & BF16 / 0. \\
EMA & Causal-rCM's iteration-dependent EMA rule with
\texttt{ema.rate=0.1}. \\
Evaluation weights & Student \texttt{net\_ema}, exported in BF16. \\
LoRA parameters & Rank 64; alpha 64; dropout 0;
zero-output initialization. \\
LoRA target layers & Q, K, V, and output projections in self- and
cross-attention; both feed-forward linear projections. \\
LoRA / full-parameter update scope & LoRA: fake-score adapters only.
Direct-Full: all fake-score parameters, without LoRA.
Both: all student parameters. \\
\end{longtable}
\endgroup

\section{Evaluation Protocol and Checkpoint Selection}
\label{app:evaluation}

\paragraph{Evaluation coverage.}
We evaluate the full VBench prompt set across all 16 dimensions, using
three seed indices (0, 1, and 2). All models use the same prompts, seed
indices, and sampling configuration. Videos contain 81 frames
at 16 FPS, with $832\times480$ resolution, four denoising steps, and chunks
of three latent frames.

\paragraph{Score aggregation.}
Following the existing evaluation pipeline, each raw dimension score is
normalized by its prescribed range and clipped to $[0,1]$. Quality dimensions
have unit weight except dynamic degree, whose weight is 0.5, giving a total
group weight of 6.5. The nine semantic dimensions have unit weight. The total
score is $0.8$ times the quality score plus $0.2$ times the semantic score.
Tables~\ref{tab:context-quality} and~\ref{tab:context-dimensions} multiply
their respective aggregate and raw scores by 100. The raw dimension scores
in Table~\ref{tab:context-dimensions} are reported before the normalization
used for aggregation.

\paragraph{Quantitative checkpoints.}
The clean-history reference is trained under the clean-history policy using
the same student initialization, text data, and DMD training setup as the
other two policies, with rank-64 fake-score LoRA and full-parameter student
updates. It uses four-step sampling with clean-context refresh.
For Table~\ref{tab:context-quality}, the selected same-level
and progressive fake-score LoRA checkpoints are 2250 and 3150, respectively.
They maximize the aggregate score among each route's completed evaluations
under the above protocol. Checkpoint numbers denote stored iteration
identifiers; equal iteration counts are not imposed across routes.

\section{Complete Dimension-Level Results}
\label{app:dimension-results}

Table~\ref{tab:context-dimensions} reports all 16 VBench dimensions for the
same three model configurations as Table~\ref{tab:context-quality}, under
the evaluation protocol in Appendix~\ref{app:evaluation}.

\begin{table}[!htbp]
    \centering
    \caption{Complete VBench dimension-level results on the full prompt set.
    Values are raw scores multiplied by 100, before normalization for
    aggregation. Higher is better in each dimension; the best result in each
    row is bold, including ties.}
    \label{tab:context-dimensions}
    \begingroup
    \setlength{\tabcolsep}{8pt}
    \renewcommand{\arraystretch}{1.12}
    \begin{tabular}{@{}lrrr@{}}
        \toprule
        Dimension & Clean & Same-level & Progressive \\
        \midrule
        \multicolumn{4}{@{}l}{\emph{Quality dimensions}} \\
        \addlinespace[2pt]
        Subject consistency & 95.196 & 95.801 & \textbf{96.286} \\
        Background consistency & \textbf{95.489} & 95.212 & 95.406 \\
        Temporal flickering & \textbf{98.511} & 98.424 & 98.453 \\
        Motion smoothness & 98.070 & 97.886 & \textbf{98.089} \\
        Aesthetic quality & 65.285 & \textbf{66.352} & 66.082 \\
        Imaging quality & 68.946 & 69.347 & \textbf{70.049} \\
        Dynamic degree & 77.778 & 84.722 & \textbf{86.111} \\
        \midrule
        \multicolumn{4}{@{}l}{\emph{Semantic dimensions}} \\
        \addlinespace[2pt]
        Object class & 95.417 & 93.646 & \textbf{96.979} \\
        Multiple objects & 78.733 & 86.458 & \textbf{87.153} \\
        Human action & 98.611 & \textbf{100.000} & \textbf{100.000} \\
        Color & 85.049 & 86.660 & \textbf{89.688} \\
        Spatial relationship & 80.963 & \textbf{81.961} & 81.504 \\
        Scene & 65.625 & \textbf{68.576} & 64.670 \\
        Appearance style & \textbf{22.543} & 21.846 & 21.936 \\
        Temporal style & 26.871 & \textbf{26.903} & 26.543 \\
        Overall consistency & 26.842 & \textbf{26.884} & 26.822 \\
        \bottomrule
    \end{tabular}
    \endgroup
\end{table}

\end{document}